\documentclass{article}
\usepackage{spconf,amsmath,graphicx,hyperref}
\usepackage{bm}
\newtheorem{remark}{Remark}
\usepackage{multirow}
\usepackage{amssymb}
\usepackage{xurl}
\title{Fed-GAME: Personalized Federated Learning with Graph Attention Mixture-of-Experts For Time-Series Forecasting}
\name{Yi Li$^{1}$,  Han Liu$^{2}$,  Mingfeng Fan$^{3}$,  Guo Chen$^{4}$, Chaojie Li$^{5*}$, Biplab Sikdar$^{2}$}
  
  \address{$^{1}$ School of Automation, Central South University \\
      $^{2}$Department of Electrical and Computer Engineering, National University of Singapore\\
      $^{3}$Department of Mechanical Engineering, National University of Singapore\\
      $^{4}$  School of Electrical Engineering and Telecommunications, University of New South Wales\\
      $^{5}$  Department of Electrical Engineering, City University of Hong Kong
      \thanks{$^{*}$Corresponding author: Chaojie Li ( chaojili@cityu.edu.hk)}
      }

\begin{document}
\topmargin=0mm
%\ninept
%
\maketitle
\begin{abstract}

Federated learning (FL) on graphs shows promise for distributed time-series forecasting. Yet, existing methods rely on static topologies and struggle with client heterogeneity. We propose Fed-GAME, a framework that models personalized aggregation as message passing over a learnable dynamic implicit graph. The core is a decoupled parameter difference-based update protocol, where clients transmit parameter differences between their fine-tuned private model  
and a shared global model. On the server, these differences 
are decomposed into two streams: (1) averaged difference used to updating the global model for consensus (2) the selective  difference
fed into a novel Graph Attention Mixture-of-Experts (GAME) aggregator for fine-grained personalization. In this aggregator,  shared experts provide scoring signals while personalized gates adaptively weight selective updates to support personalized aggregation. Experiments on two real-world electric vehicle charging datasets demonstrate that Fed-GAME outperforms state-of-the-art personalized FL baselines.%, validating the effectiveness of Fed-GAME.
\end{abstract} 
\section{Introduction}
\label{sec:intro}

Federated Learning (FL) enables collaborative training over decentralized data \cite{mcmahan2017communication}, but its performance is often degraded by statistical heterogeneity. This problem is particularly severe in distributed time-series forecasting tasks such as EV charging demand prediction, where each client exhibits distinct temporal patterns and seasonality. 
Personalized FL (PFL) addresses this challenge by tailoring models to local characteristics through techniques such as proximal regularization, clustering, or multi-task learning %\cite{li2020federated,tan2022towards,fallah2020personalized,wang2022accelerating,t2020personalized,10446879}. 
\cite{li2020federated,tan2022towards,wang2022accelerating,t2020personalized,10446879}.
However, many of these approaches emphasize local adaptation (e.g., extended local training or personalized regularization) to improve client performance, while overlooking a core aspect of FL: the aggregation of knowledge across clients.
%To improve personalization, recent works introduce Federated Graph Learning (FGL), 
Since aggregation directly impacts the quality of personalization under heterogeneous data, recent works propose Federated Graph Learning (FGL), enabling clients to collaborate more selectively with similar peers rather than relying on uniform aggregation. \cite{chen2022personalized, meng2021cross,liu2024federated,zhou2025personalized}. For instance, \cite{chen2022personalized} designed bilevel optimization schemes to train both local models and the GNN model with two types of objective functions.  \cite{meng2021cross} leveraged geographical proximity as client graph, to guide the model aggregation process. However, existing FGL and other advanced PFL approaches face two fundamental challenges. 

\textbf{ (1) Imprecise Client Graphs}: 
%Most approaches rely on predefined graphs (e.g., based on geography), which may not accurately reflect the true task-specific relationships between clients, or infer dynamic graphs from full model updates \cite{zhou2025personalized, pmlr, li2024federated}. While these method are more adaptive, typically requires clients to transmit their full model parameters for similarity calculation. This still requires clients to transmit their entire models, a communication-intensive process, and the resulting graph is often based on a holistic similarity measure rather than a fine-grained, personalized one.
% Most FGL approaches rely on predefined, static graphs (e.g., based on geography) \cite{zhou2025personalized, pmlr}, which may not reflect the true task-specific relationships between clients. While some works infer dynamic graphs from full model updates \cite{li2024federated}, this still requires clients to transmit their entire models, a communication-intensive process, and the resulting graph is often based on a holistic similarity measure rather than a fine-grained, personalized one.
Existing FGL methods often rely on predefined, static graphs (e.g., based on geography) \cite{zhou2025personalized, pmlr}, which may not capture the true task-specific relationships between clients. Some studies attempt to infer dynamic graphs from full model updates \cite{li2024federated}, but this requires clients to transmit their entire models, and the resulting graphs are typically based on coarse-grained, global similarity measures rather than fine-grained, personalized ones.

%Suboptimal Aggregation of Mixed-Information Updates}: Most PFL frameworks rely on the exchange of a single, mixed-information update from each client. This single update vector conflates a client's contribution to the global consensus with its unique personalization needs. Applying a uniform aggregation strategy (e.g., weighted averaging or a single attention mechanism) to this mixed signal is inherently suboptimal. It can lead to personalization signals from some clients detrimentally affecting the shared model for others, or conversely, the drive for consensus can suppress valuable, distinct personalization signals.
\textbf{(2) Suboptimal Aggregation of Mixed Updates}: 
%Most PFL frameworks exchange a single update vector per client, conflating global consensus with client-specific personalization. However, the knowledge learned by a client should be decoupled into two categories: the general knowledge across all clients and client-specific knowledge for this client\cite{Wu_2023_ICCV}. Recent study\cite{10.1145/3664647.3681588} uses parameter additive decomposition to extract parameters of general knowledge and locally retain parameters to learn client-specific knowledge. However, these method does not take into account the relationship between clients.
% The key goal of PFL is to separate client-specific knowledge from general knowledge, since indiscriminate aggregation can degrade performance under client heterogeneity \cite{Wu_2023_ICCV}. In conventional FL, server-side aggregation (e.g., FedAvg) combines heterogeneous client updates that simultaneously contain both global and client-specific knowledge. Such naive averaging is suboptimal: the global patterns are diluted while the private patterns are erroneously merged, leading to a model that generalizes poorly and fails to adapt to individual clients.
% Recent work \cite{10.1145/3664647.3681588} addresses this by using parameter additive decomposition, extracting global parameters for knowledge sharing while retaining local parameters for personalization. However, this decomposition neglects the relationships among clients.
In conventional PFL frameworks, each client only exchanges a single update vector, which conflates global consensus with client-specific personalization \cite{Wu_2023_ICCV}. 
%This update simultaneously encodes both the client’s contribution to the global consensus and its unique personalized preferences. Naively averaging these mixed signals, 
%By simply averaging these updates, as in FedAvg, will weakens general knowledge and obscures personalized signals. The resulting global model may fail to serve different client effectively. 
For example, FedAvg \cite{mcmahan2017communication} simply averages the updates, which weakens general knowledge and obscures personalized signals. The resulting global model may fail to serve different client effectively. Recent parameter decomposition techniques \cite{10.1145/3664647.3681588} addresses this by using parameter additive decomposition, extracting global parameters for knowledge sharing while retaining local parameters for personalization. However, this decomposition neglects the rich inter-client relationships that are crucial for truly effective personalization.

To track these challenges, we propose personalized \underline{F}ederated  Learning with \underline{G}raph \underline{A}ttention \underline{M}ixture-of-\underline{E}xperts (Fed-GAME), a novel PFL framework for time-series forecasting. Our contributions are:
% 1. We separate global consensus from personalization by transmitting (i) full parameters of the shared global model ($M_B$) for robust convergence, and (ii) selective sparse parameter differences ($M_B - M_A$) for personalization. This design reduces communication overhead while providing fine-grained personalization signals, directly addressing the challenge of mixed updates.
(1) We separate global consensus from personalization by transmitting only the parameter differences between each client’s fine-tuned private model ($M_A$) and the shared global model ($M_B$). This design enables fine-grained personalized updates and  maintain robust global consensus. %, directly addressing the challenge of mixed updates in heterogeneous clients.
(2) We propose a server-side GAME aggregator that learns dynamic, client-specific aggregation strategies from the transmitted updates, allowing GAME capture task-specific relationships in a flexible, data-driven manner. % This directly addresses the challenge of imprecise client graphs by capturing task-specific relationships in a flexible, data-driven manner.
(3) We design a similarity-based meta-loss to train the aggregator in server without direct access to client data.

%\vspace{-0.2cm}
\section{The Fed-GAME Framework}
\label{sec:format}

\subsection{Problem Formulation}
The goal of this paper is personalized time-series forecasting for $N$ clients in federated setting. Each client $i\in \{1,\cdots, N\}$ 
possesses a local private dataset $D_{i} = \{(\textbf{X}_i,\textbf{Y}_i)\}$, 
where $\textbf{X}_i=\{X_{i}^{t+1},\cdots,X_{i}^{t+h}\}$ represents the historical sequence of length $h$, 
and $\textbf{Y}_i=\{y_{i}^{t+h+1},\cdots,y_{i}^{t+h+p}\}$ is the $p$-step corresponding future sequence to be predicted. 
The forecast series of $p$-steps of $i$-th client is denoted as $\hat{\textbf{Y}}^{p}_{i}=\{\hat{y}_{i}^{t+h+1},\cdots, \hat{y}_{i}^{t+h+p}\}$.
Due to data heterogeneity, %the data distributions vary across clients. Therefore, 
instead of learning a single global model, Fed-GAME aims to learn a unique set of model parameters $\bm{w}_i$ for each client. The objective is to minimize the aggregated loss over all clients: $
    \min_{\bm{w}}\sum_{i\in N}  \lambda_i f_i(\bm{w};\mathcal{D}_i)$,
where $N$ denotes the number of clients, $f_i(\cdot)$ is the forecasting loss for client $i$, $\lambda_i$ reflects the dataset weight, $\bm{w}$ is the model parameters to be optimized.% and $[N]$ represents the total clients in the federated learning.

%\vspace{-0.2cm}
\subsection{Fed-Game Framework}
%To address the aforementioned challenges, % of suboptimal aggregation and high communication costs in PFL for time-series,
To realize personalized time-series forecasting,
 we propose Fed-GAME. %a novel personalized federated learning framework.
%as illustrated in Fig. \ref{fig:framework}.
Fed-GAME involves the following steps, each labeled by stage number in Fig. \ref{fig:framework} for clarity:
\begin{figure}
    \centering
    \includegraphics[width=1\linewidth]{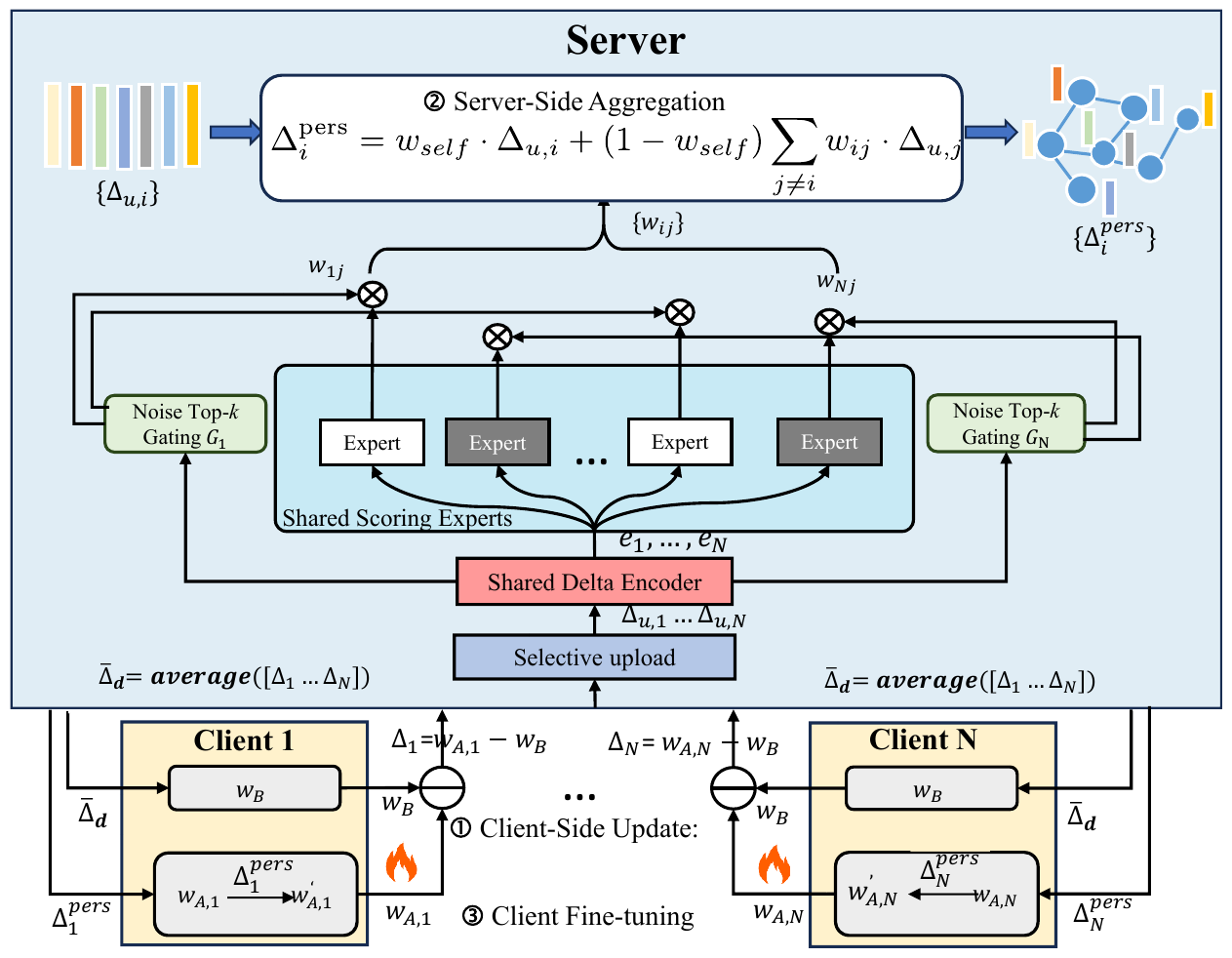}
    \caption{The Framework of Fed-GAME}
    \label{fig:framework}
\end{figure}

\textbf{(1) Client-Side Update}: 
%the server dispatches the global delta update $\overline{\Delta}_{d}^{t}$ from the previous round $t$ , and a personalized delta update $\Delta_{d, i}^{\text{pers}, t}$ to each client $i$. The global delta $\overline{\Delta}_d^{t}$  represents the averaged knowledge from the previous round. The personalized update $\Delta_{d, i}^{\text{pers}, t}$ denotes the personal knowledge. 
Each client $i$ maintains its own persistent private model ($M_A$) with parameters $w^{\tau}_{A,i}$  
and a local copy of the shared global model ($M_B$) with parameters $w^{\tau}_B$ in round $\tau$. 
After fine-tuning $M_A$ on its local data in round $\tau$, client $i$ computes the parameter difference relative to the global model:
$\Delta^{\tau}_i=w^{\tau}_{A,i}-w^{\tau}_B$.
%, where $w^{\tau}_{A,i}$ is the parameter of locally fine-tune private model. 
The client then uploads  $\Delta^{\tau}_{i}$ to the server. 

\textbf{(2) Server-Side Aggregation}:  
On the server, these difference are used for two purposes:
i. Global consensus update: 
%The averaged delta $\overline{\Delta}_d^{t} = \frac{1}{N}\sum_{i}\Delta^{t}_{i}$, updates the shared global model as $M^{t+1}_{B} = M^{t}_{B}+ \eta\cdot \overline{\Delta}^{t}_d$.
To maintain a shared knowledge base, the server averages client  $\Delta^{\tau}_{i}$ and applies them to the global model:
$\overline{\Delta}_d^{\tau} = \frac{1}{N}\sum_{i}\Delta^{\tau}_{i}$,
$w^{\tau+1}_{B} = w^{\tau}_{B}+ \eta\cdot \overline{\Delta}^{\tau}_d$. Here, $\eta$ is the server learning rate controlling the step size of the global update.
ii. Personalized aggregation: 
For communication efficiency, only selective components of $\Delta_{i}^{\tau}$ (denoted as $\Delta_{u,i}^{\tau}$ ) are utilized for personalized aggregation. Here, we restrict to select the final MLP layers, as they are widely recognized to capture high-level transferable features, whereas lower layers remain client-specific.
Then, each client’s selective difference $\Delta^{\tau}_{u,i}$ is encoded by a shared Encoder (constructed by Linear layer) into a low-dimensional embedding $\mathbf{e}^{\tau}_i = \textit{Encoder}(\Delta^{\tau}_{u,i})$
to stabilize training and reduce dimensionality for efficient attention computation.
These embeddings are then fed into a GAME aggregator, where shared experts model inter-client relevance and personalized gates adaptively weight them. The resulting scores ${w_{ij}^{\tau}}$ serve as aggregation weights for $M_A$ update, yielding the personalized update $\Delta_{i}^{\text{pers}, \tau}$ as:
\begin{equation}
   \Delta_{i}^{\text{pers}, \tau} = w_{self}\cdot \Delta_{u,i}^{\tau}+ (1-w_{self})\sum_{j\neq i}  w^{\tau}_{ij}\cdot \Delta_{u,j}^{\tau},
   \label{message-passing}
\end{equation}
where $w_{self}$ controls the balance between a client's local updates and the aggregated knowledge from its peers. This corresponds to graph-based message passing on the client interaction graph, where edges encode similarity in final-layer updates. The computation of the graph attention weights $\{w^{\tau}_{ij}\}$ is detailed in Section 2.3.

% \textbf{3). Client Fine-tuning}: Clients first incorporate the personalized delta into their private model, then perform local fine-tuning, aligning the global representation with personalized objectives: $M^{t+1}_{A,i} = M^{t}_{A,i}+ \gamma\cdot \Delta_d^{\text{pers,i}, t}$. Then, fine-tune $M^{t+1}_{A,i}$ on local data $M^{t+1'}_{A,i}$, the next round delta is calculated as $\Delta^{t+1}_i=M^{t+1'}_{A,i}-M^{t+1}_B$.

\textbf{(3) Client Fine-tuning}: %Each client updates its private model with the personalized delta and then fine-tunes on local data:
Each client applies the personalized $\Delta^{pers,\tau}_{i}$ and fine-tunes on local data:
\begin{equation}
w_{A,i}^{\tau+1} = \text{FineTune}(w_{A,i}^{\tau’}) = \text{FineTune}(w_{A,i}^{\tau} + \gamma \cdot \Delta_{i}^{\text{pers},\tau}),
\end{equation}
 $\gamma$ denotes the learning rate controlling the step size of the personalized update.
After fine-tuning, the next-round $\Delta_i^{\tau+1}$ is computed as $\Delta_i^{\tau+1} = w_{A,i}^{\tau+1} - w_B^{\tau+1}$ for next round upload.

%\vspace{-0.2cm}
\subsection{GAME Aggregator  for Aggregation Weights}
%In our study, we use the Mixture-of-Expert (MoE) \cite{moe} structure to handel different attention between clients. Our MoE aggregator comprises: A set of shared scoring experts that learn to evaluate the relevance between pairs of client embeddings ($\textbf{e}_j$). The number of experts is set as $M$, $N$ represents the total number of samples. Each expert $E^{j}$ is a neural network designed to capture client-specific time-series features and then give a score to 
% The core of our server-side personalization is the learnable aggregator, which is based on the Mixture-of-Expert (MoE) \cite{moe} structure. %We reframe the knowledge sharing process as a message passing scheme on a graph and proposed Graph Attention Mixture-of-Expert (GAME) aggregator. 
% For each round (omitting the superscript $t$ for clarity),
% we construct a fully connected implicit graph on the server where each uploaded update $\Delta_{u,i}$ is represented as a node. The low dimensional embedding $\textbf{e}_{i}$ of this update serves as the dynamic node feature, and the raw high dimensional $\Delta_{u,i}$ acts as the message to be propagated on the graph. Unlike approaches that rely on a fixed adjacency matrix, our proposed MoE module learns to compute personalized graph attention coefficients $\{w_{ij}\}_{j\neq i}$, which dynamically determine the edge weights for each target node. These coefficients are then used to aggregate the messages from neighbor nodes. This mechanism is composed of shared scoring experts and personalized client gates.

The core of our server-side personalization is a learnable Mixture-of-Experts (MoE) aggregator. In each round (superscript 
$\tau$ omitted for clarity), we construct a fully connected implicit graph where each uploaded update  $\Delta_{u,i}$
 is treated as a node. Its low-dimensional embedding $\textbf{e}_{i}$ serves as the dynamic node feature, while the raw update $\Delta_{u,i}$ is propagated as the message.
Unlike approaches with fixed adjacency matrices, our aggregator learns personalized graph attention coefficients $\{w_{ij}\}_{j\neq i}$, which adaptively weight messages (raw update $\Delta_{u,i}$) for each client. This personalized attention mechanism is realized through two components: shared scoring experts, which capture inter-client relevance, and  personalized gating modules, which adaptively weight expert outputs for each client.

\textbf{(1) Shared Scoring Experts}: We employ $M$ shared scoring experts $\{E_{k}\}^{M}_{k=1}$ (construct by linear networks). The role of these experts is to evaluate the relevance of neighbor node $j$'s message to a target client $i$. The scalar score $s_{ijk}$ is calculated as:
%\begin{equation}
    $s_{ijk} = E_k([\bm{e}_j,\bm{e}_i])$. This score represents the assessment of a specific expert regarding the relevance between each pair of clients.
%\end{equation}

\textbf{(2) Personalized Noisy Top-$k$ Gating}: 
Each client $i$ is equipped with a personalized gate $G_i$, which assigns combination weights over experts. The gate generates logits $H_i(e_i) = (e_i \cdot w_{gi}) + \text{SN}(\text{softplus}((e_i \cdot w_{\text{noise}i})))$, where $w_{gi}$ and $w_{\text{noise}i}$ are learnable parameters, SN adds Gaussian noise for exploration, and $\text{softplus}(z) = \log(1+e^z)$. Applying Top-$k$ selection followed by softmax produces a sparse weight vector: $c_i = G_i(e_i) = \text{Softmax}(\text{Top-}k(H_i(e_i), k))$, which combines expert scores into a single relevance value: $v_{ij} = c_i^\top s_{ij}, s_{ij} = [s_{ij1}, \dots, s_{ijM}]$. Graph attention coefficients are then obtained via temperature-scaled softmax: $w_{ij} = \exp(v_{ij}/T) / \sum_{l\neq i} \exp(v_{il}/T)$, where $T$ is the temperature. Thus, $w_{ij}$ defines the edge weights of a dynamic, client-centric graph for each round, computed via weighted message passing in Eq. (\ref{message-passing}); $w_{ij}$ can also be interpreted as personalized weight. details can be found in \cite{11180413}.

\vspace{-0.2cm}
\begin{remark}
Compared with standard graph attention aggregators (e.g., GAT), Fed-GAME eliminates reliance on static topology and adopts a MoE paradigm, decomposing the single attention function into multiple shared experts and personalized routers, enabling both multi-expert relationship modeling and client-specific aggregation strategies.
\end{remark}
\subsection{Training Objectives}
The Fed-GAME framework employs a decoupled, two-level optimization strategy.
\textbf{Client-Side}: Each client 
$i$  optimizes its private model $M_A$  by minimizing a local objective function, which combines the forecasting loss on its local data with a proximal term from FedProx \cite{li2020federated} to regularize local training and mitigate drift from the global model $M_B$. The local training of client $i$ is formulated as: $\min_{w_{A}} \left( \mathcal{L}_{\text{task}}^i(w_{A}) + \frac{\mu}{2} \left| w_{A} - w_{B} \right|_2^2 \right)$.
% \begin{equation}
% \min_{w_{A}} \left( \mathcal{L}_{\text{task}}^i(w_{A}) + \frac{\mu}{2} \left| w_{A} - w_{B} \right|_2^2 \right)
% \end{equation}
\textbf{Server-Side}: The server trains its learnable aggregation module (Encoder and MoE) using a similarity-based meta-loss $L_{meta}$, 
which is defined as a weighted combination of a Mean Squared Error (L2) loss and a cosine dissimilarity loss:
\begin{align}
\label{eq:meta_loss}
\mathcal{L}_{\text{meta}}^{i}(w_{\text{enc}}, w_{\text{moe}}) = & \alpha \left| \Delta^{pers}_{i} - \Delta_{u,i} \right|_2^2 \nonumber \\
& + \beta  \left(1 - \frac{\Delta^{pers}_{i} \Delta_{u, i}}{|\Delta_{i}^{pers}| |\Delta_{u, i}|} \right)
\end{align}
where $\alpha$ and $\beta$  are balancing hyperparameters, $w_{\text{enc}}$ and $w_{\text{moe}}$ denote the parameters of the Encoder and MoE.
%This loss is computed on the server and aims to align the MoE-generated personalized updates with the clients' own raw updates. 
This meta-loss encourages the MoE to generate personalized updates that align with clients’ own raw parameter difference, improving personalization while maintaining global consistency.
%It minimizes a combination of the L2 distance and cosine dissimilarity between the expected personalized delta and the raw client delta for the model's MLP layers. The server's modules are updated by taking a gradient step on this meta-loss, allowing them to learn a content-aware personalization strategy without direct access to client data.
\subsection{Communication Cost Analysis}
Communication overhead is a critical bottleneck in FL. Therefore, we analyze the communication cost per round of Fed-GAME and compare it with standard FL baselines like FedAvg and FedProx that exchange full model parameters. Let 
%$N$ be the number of clients participating in a round, and 
$\theta$  be the total number of parameters in the model. For baselines, both upload and download involve full parameters, and the total communication cost is $C_{baseline}= 2N|\theta|$. 
For Fed-GAME, let $r$ be the fraction of parameters selected for communication $0<r<1$. 
The total upstream cost is $N\cdot |\theta|$. For the downstream, the server sends a global parameter difference ($\overline{\Delta}_d$) and a  personalized parameter difference ($\Delta_i^{\text{pers}}$). The total communication cost is
$C_{\text{Fed-GAME}} = 2N|\theta| + N \cdot r|\theta| = (2+r)N|\theta|$.
%\end{equation}
%For a large $N$, the cost can be approximated as:
%\begin{equation}
%\label{eq:cost_fedgame_approx_final}
%$C_{\text{Fed-GAME}} \approx 2Nk|\theta|$.
%\end{equation}
By comparing this with the baseline, we derive the communication cost ratio:
% \begin{equation}
% \label{eq:ratio_final}
$\text{Cost Ratio} = \frac{C_{\text{Fed-GAME}}}{C_{\text{Baseline}}} \approx \frac{(2+r)N|\theta|}{2N|\theta|} = 1+\frac{r}{2}$.
%\end{equation}
The personalized head is tiny ($r<1\%$), so, the overhead is negligible while enabling stronger server-side MoE aggregation.

%Then the expert scores for $j$, $\textbf{s}_{ij}=[s_{ij1},\cdots,s_{ijM}]^{T}$ are combined using the client $i$'s gate to produce a single value score : $v_{ij}=\textbf{c}^{T}_{i}\textbf{s}_{ij}$. 
% \begin{equation}
%   Top\\-k(v,k) = 
%   \begin{cases} 
%   v_i, & \text{if } v_i \text{ is among the top-} k \text{ values in } v, \\
%   0, & \text{otherwise}.
%   \end{cases}
% \end{equation}

% Each client have a personalized \textit{gating mechanism} for each client $i$ , which takes its own embedding $\textbf{e}_i$ as input and learns how to combine the scores from the shared experts. The final output of the MoE is a set of personalized graph attention coefficients $\{w_{i,j}\}_{i\neq j}$. These learned weights $w_ij$ are then used to compute a personalized "others contribution" by taking a weighted average of the raw deltas from neighbor clients. This is combined with the client's own raw delta (controlled by a self-weighted) to form the final personalized update, which is then sent back to the client. The parameters of the Delta Encoder and MoE are trained on the server via a similarity-based meta-loss.

\vspace{-0.2cm}
\section{Experimental Results}
\label{sec:pagestyle}
We conduct case studies on two real-world datasets, Palo
Alto \cite{EVdata} and Shenzhen \cite{zhou2021shenzhen}, which capture small and large scale non-IID spatial-temporal demand patterns. The Palo Alto dataset contains 8 charging stations with demand (kWh) recorded at 5-minute intervals, while the Shenzhen dataset includes 247 stations with 30-minute resolution. %For details, please refer to the references.

We compare Fed-GAME with SOTA,
including FedAvg \cite{wang2022accelerating}, FedProx \cite{li2020federated}, pFedMe \cite{t2020personalized}, PAG-FedAVG \cite{qu2024physics} and GCRN-FedAvg \cite{fahim2024forecasting}, where FedProx and pFedMe are PFL, while PAG-FedAVG and GCRN-FedAVG are FGL. We also consider No\_FL which trains a forecasting model locally on each client without any communication. For the uncertainty of the charging demand, we use quantile regression for time-series forecasting \cite{11075774}.
Model performance is evaluated using three standard criteria: Quantile Score (QS), Mean Interval Length (MIL), and Interval Coverage Percentage (ICP), summarized in Table \ref{Evaluation Metrics}.
\begin{table}[]
\setlength{\abovecaptionskip}{0cm}
\setlength{\belowcaptionskip}{-0.6cm}
\centering
\vspace{-0.4cm}
\caption{Evaluation Metrics}
\label{Evaluation Metrics}
\scalebox{0.60}{
\begin{tabular}{c}
\hline \hline
QS =
\(
\begin{cases}
    \frac{1}{n}\sum^{n}_{i=1}(1-q)\|y^{t}_{i}-\hat{y}^{t}_{i,q}\|, & y^{t}_{i}<\hat{y}^{t}_{i,q} \\
    \frac{1}{n}\sum^{n}_{i=1}q\|y^{t}_{i}-\hat{y}^{t}_{i,q}\|, & y^{t}_{i}\geq\hat{y}^{t}_{i,q}
\end{cases}
\)
\\[6pt]
ICP =
\(
\frac{1}{n}\sum^{n}_{i=1}
\begin{cases}
   1, & \hat{y}^{t}_{i,q}\leq y^{t}_{i}\leq\hat{y}^{t}_{i,q^{'}} \\
   0, & \text{otherwise}
\end{cases}
\)
\\[6pt]
MIL = \(\frac{1}{n}\sum^{n}_{i=1}\|\hat{y}^{t}_{i,q}-\hat{y}^{t}_{i,q^{'}}\|\)
\\
\hline \hline
\end{tabular}}
\end{table}
%The experiments are set by 30 minutes (horizon 6), 60
% minutes (horizon 12) ahead forecasting based on 3-hours historical input data from the Palo
% Alto and 3 hours (horizon 6) and 6 hours (horizon 12) aheah forecasting based on 6-hours historical input data from the ShenZhen datasets. 
% For the Palo Alto dataset, forecasts are conducted 30 minutes (horizon 6) and 60 minutes (horizon 12) ahead based on 3 hours of historical input. For the Shenzhen dataset, forecasts are conducted 3 hours (horizon 6) and 6 hours (horizon 12) ahead based on 6 hours of historical input.
% In principle, any prediction model can be used. However, to reflect practical device constraints, we employ a two-layer LSTM+MLP model, with LSTM hidden dimensions of 256 and 128. For Fed-GAME, the learning rate  is set as 0.0005, with $w_{self} = 0.6$, $\mu = 0.2$, $\alpha = \beta= 0.5$. 4 shared scoring experts, and top-$k = 2$. Hyperparameters were determined via preliminary experiments. To evaluate probabilistic forecasting, $q$ is set to $\{0.1, 0.5, 0.9\}$, allowing assessment across confidence intervals (ICP should approach 0.8, while MIL should be minimized). All experiments are implemented in Python 3.10 on an NVIDIA A100 GPU.
For the Palo Alto dataset, we use 3 hours historical data to predict the next 30 minutes (horizon 6) and 60 minutes (horizon 12) charging demand. For Shenzhen, 6 hours historical data are used to predict 3 hours (horizon 6) and 6 hours (horizon 12) charging demand. In principle, any prediction model can be used. However, to reflect practical device constraints, we employ a two-layer LSTM+MLP model, with LSTM hidden dimensions of 256 and 128. For Fed-GAME, the learning rate  is set as 0.0005, with $w_{self} = 0.6$, $\mu = 0.2$, $\alpha = \beta= 0.5$. Four shared scoring experts, and Top-$k = 2$ was used. Hyperparameters were determined via preliminary experiments. To evaluate probabilistic forecasting, $q=\{0.1, 0.5, 0.9\}$, %allowing assessment across confidence intervals (
ICP should approach 0.8, while QS and MIL should be minimized. All experiments are implemented in Python 3.10 on an NVIDIA A100 GPU.

%\subsection{Predictive Performance Compare with Baseline}
\vspace{-0.2cm}
\subsection{Comparison Analysis}
The forecasting performance results of Palo Alto and Shenzhen are shown %in Table. \ref{QS_1} and \ref{QS_2}, respectively.
in Table. \ref{QS_merge}.
On the Palo Alto dataset, Fed-GAME achieves the best average
QS and ICP scores. While our method does not achieve the best MIL performance as we upload only parameter differences rather than full parameters, it still surpasses the baselines in QS and ICP. This confirms that Fed-GAME’s aggregation strategy can effectively harness client knowledge without direct data sharing. %For demonstration, the 6-step ahead
%prediction results of the Fed-GAME for different stations are drawn,
%as shown in Fig. 3. 
On the Shenzhen dataset,
with 200 stations (47 unseen during training), Fed-GAME improves
Step-6 performance by 54.23\% (QS) and 43.87\% (ICP) over
No\_FL. For Step-12, improvements are 45.14\% and 68.93\%,
demonstrating strong generalization under non-IID and partial
client participation settings. These results highlight its strength in long-term forecasting and confirm Fed-GAME’s effectiveness in both accuracy and generalization across diverse deployment scenarios.
\subsection{Performance of the GAME Aggregator}
We evaluate the effectiveness of GAME aggregator against two strong graph-based baselines: GraphSAGE \cite{Graphsage}, which uses a fixed aggregation, and GAT \cite{GAT} that learns attention weights between nodes. In this ablation, only the server-side aggregation module is replaced. The results are shown in Table. \ref{ablation_3}. The experimental results demonstrate the superiority of the GAME aggregator. On both datasets, Fed-GAME significantly outperforms both GraphSAGE and GAT on QS and MIL. This performance demonstrates the effectivenes of GAME. While GAT computes a single attention score between nodes, our GAME aggregator utilizes a set of scoring experts to evaluate client relationships from multiple perspectives. The personalized noisy Top-$k$ router learns a sparse, adaptive strategy to combine these expert's judgment  for each client. This approach allows GAME to learn effective content-aware aggregation function, resulting in better forecasting accuracy.
\begin{table}[t]
\setlength{\abovecaptionskip}{-0.2cm}
\centering
\caption{Performance comparison of different FL forecasting models on Palo Alto and Shenzhen datasets}
\label{QS_merge}
\scalebox{0.6}{
\begin{tabular}{l l l c c c| c c c}
\hline \hline
 & Method & Aggregation 
 & \multicolumn{3}{c}{Palo Alto} 
 & \multicolumn{3}{c}{Shenzhen} \\
\cline{4-9}
 &  & Method 
 & QS & MIL & ICP 
 & QS & MIL & ICP \\
\hline

\multirow{7}{*}{Step 6}
& No\_FL & --- 
& 1.5228 & 4.7110 & 0.7585 
& 18.6698 & 39.8802 & 0.6988 \\

& LSTM-FedAvg & Average
& 1.4846 & 4.5517 & 0.7390
& 10.4788 & 10.1150 & 0.7381 \\

& FedProx & Average
& 1.5378 & 4.7564 & 0.6571
& 9.8572 & \textbf{6.3500} & 0.6587 \\

& pFedMe & Average
& 1.7047 & 5.0307 & 0.6677
& 10.5375 & 9.5557 & 0.7176 \\

& PAG-FedAvg & Average
& 1.5304 & \textbf{4.2462} & 0.7534
& 13.7907 & 13.0350 & 0.4305 \\

& GCRN-FedAvg & Average
& 1.4219 & 4.2518 & 0.7448
& 9.9119 & 9.7647 & 0.7002 \\

& Fed-GAME & GAME
& \textbf{1.4067} & 4.4833 & \textbf{0.7827}
& \textbf{8.5461} & 9.8635 & \textbf{0.7432} \\

\hline
\multirow{7}{*}{Step 12}
& No\_FL & ---
& 2.4770 & 7.0310 & 0.6726
& 19.1557 & 35.0342 & 0.6310 \\

& LSTM-FedAvg & FedAvg
& 1.8118 & 5.4041 & 0.6633
& 18.4644 & 14.2590 & 0.7351 \\

& FedProx & FedAvg
& 2.1077 & 5.4319 & 0.6662
& 11.8104 & 12.4952 & 0.7260 \\

& pFedMe & FedAvg
& 2.1397 & 5.0307 & 0.6963
& 12.7760 & 13.5238 & 0.7077 \\

& PAG-FedAvg & FedAvg
& 1.8843 & 5.0488 & 0.6866
& 12.7842 & \textbf{13.3762} & 0.6907 \\

& GCRN-FedAvg & FedAvg
& 1.6297 & 5.1615 & 0.7277
& 9.3349 & 17.9791 & 0.7431 \\

& Fed-GAME & GAME
& \textbf{1.6270} & \textbf{5.0027} & \textbf{0.7569}
& \textbf{10.5097} & 15.5639 & \textbf{0.7475} \\

\hline \hline
\end{tabular}}
\end{table}
\begin{table}[]
\setlength{\abovecaptionskip}{0cm}
\centering
\vspace{-0.4cm}
	\caption{QS, MIL and ICP with different Aggregation Mechanism on different dataset (Prediction Step=12)}
        \label{ablation_3}
        \scalebox{0.6}{
	\begin{tabular}{l l l l c c c}
		\hline \hline
		          &Method      &  Aggregation Method                      & QS & MIL & ICP \\ 
		\hline
		\multirow{4}{*}{Palo Alto} 
            & No\_FL & --- &2.477   &7.031  & 0.6726\\
            &Fed-GraphSAGE &GraphSAGE & 1.8878   &5.6907 &0.756 \\
            &Fed-GAT &GAT & 1.9081  &5.6808 &0.7462 \\
            &Fed-GAME &GAME &\textbf{1.627} &\textbf{5.0027} &\textbf{0.7569}\\
		\hline
		\multirow{4}{*}{Shenzhen} 
            & No\_FL & --- &19.1557  &35.0342  & 0.631\\
            &Fed-GraphSAGE &GraphSAGE & 11.9333  &19.5901 &\textbf{0.756} \\
            &Fed-GAT &GAT & 12.0329  &19.3822 &0.7392 \\
            &Fed-GAME &GAME &\textbf{10.5097} &\textbf{15.5639} &0.7475\\	
		\hline \hline
	\end{tabular}}
\end{table}
\vspace{-0.2cm}
\subsection{Communication Efficiency Evaluation}

We compare the communication cost of Fed-GAME with baselines (FedAvg, FedProx and pFedMe). For the LSTM model ($\approx 996K $ parameters), in 6-step prediction (Palo Alto/Shenzhen) task, linear head = 2,322 parameters, the fraction of parameters selected for communication $r = \frac{|\theta_{\text{linear}}|}{|\theta|} = \frac{2,322}{996,013} \approx 0.0023$, which results in only 0.115\% overhead. In 12-step prediction (Palo Alto/Shenzhen) task, linear head = 4,644 parameters, $r= \frac{4,644}{994,852} \approx 0.0047$, which increase 0.235\% overhead. Thus, Fed-GAME achieves personalization at negligible cost compared to full-parameter baselines.

\subsection{Analysis of Learned Aggregation Weights}
We further analyze the learned aggregation weights to better substantiate the proposed mechanism. The results show that the personalized weights $w_{ij}$ are not solely determined by geographic proximity; instead, clients exhibit specialization across different temporal scales, leading to structured attention patterns. As training progresses, the aggregation weights evolve from near-uniform distributions to more differentiated profiles with higher variance and lower entropy, indicating the capture of non-trivial inter-client correlations under non-IID data distributions and yielding a 54.23\% QS improvement on the Shenzhen dataset (Table. ~\ref{QS_merge}).

\vspace{-0.2cm}
\section{Conclusion}
In this paper, we proposed a novel PFL framework, Fed-GAME, for time-seires forecasting by employing GAME aggregator to learn dynamic, content-aware graph attention coefficients. Experimental results demonstrate that FED-GAME achieves state-of-the-art predictive performance on heterogeneous time-series data. For future work, we plan to further investigate adaptive strategies for parameter selection in the selective update process.
\vfill\pagebreak

% \section{REFERENCES}
% \label{sec:refs}

\bibliographystyle{IEEEbib}
\bibliography{refs}

\end{document}